\documentclass[sigconf]{acmart}

\renewcommand\footnotetextcopyrightpermission[1]{}
\AtBeginDocument{%
  }

\usepackage{booktabs}
\usepackage{multirow}
\usepackage{array}
\usepackage{algorithm}
\usepackage{algpseudocode}
\usepackage{placeins}
\usepackage{xcolor}

\definecolor{gold}{RGB}{212,175,55}
\definecolor{silver}{RGB}{160,160,160}
\definecolor{bronze}{RGB}{205,127,50}
\definecolor{bestred}{RGB}{255,59,48}
\definecolor{secondblue}{RGB}{0,122,255}

\newcommand{\best}[1]{\textcolor{bestred}{\textbf{#1}}}
\newcommand{\second}[1]{\textcolor{secondblue}{\textbf{#1}}}

\title[TC-MAF for Multimodal Industrial AD]{TC-MAF: Train-Calibrated Bounded Multi-Evidence Fusion for Multimodal Industrial Anomaly Detection}

\author{Ming Deng}
\affiliation{%
  \institution{Shanghai University}
  \city{Shanghai}
  \country{China}
}
\email{chranos@shu.edu.cn}

\author{Sijin Sun}
\affiliation{%
  \institution{National University of Singapore}
  \city{Singapore}
  \country{Singapore}
}
\email{sun.sijin@u.nus.edu}

\author{Xiaochuan Hu}
\affiliation{%
  \institution{University of Electronic Science and Technology of China}
  \city{Chengdu}
  \country{China}
}
\email{202421090208@std.uestc.edu.cn}

\author{Xing Wu} \authornote{Corresponding author.}
\affiliation{%
  \institution{Shanghai University}
  \city{Shanghai}
  \country{China}
}
\email{xingwu@shu.edu.cn}

\begin{document}

\begin{abstract}
Multimodal anomaly detection benefits from complementary RGB and 3D evidence, yet auxiliary RGB reconstruction is not equally reliable across product categories and class-wise test-time policy selection is usually unavailable. We propose TC-MAF, a base-anchored multi-evidence fusion design that combines a multimodal detector, complementary Dinomaly evidence, and a small cross-modal consistency cue under one fixed pixel-level fusion formula. A lightweight training-dispersion confidence (TDC) term scales auxiliary participation using only normal training statistics. On MVTec-3D, TC-MAF reaches 0.979 image-level AUROC and 0.990 pixel-level AUPRO, achieving the best mean results on both detection and localization among the compared multimodal methods. Systematic ablations show that the fusion structure itself is the dominant factor, while TDC provides a smaller but reproducible calibration gain over no calibration or arbitrary calibration. Additional experiments show that the same design remains effective under a pooled-statistics variant, auxiliary-branch and backbone substitutions, few-shot settings, a missing-3D setting, and cross-dataset evaluation on Eyecandies. Code is available at \url{https://anonymous.4open.science/r/TC_MAF-C3BB}.
\end{abstract}



\maketitle

\begin{figure}[t]
  \centering
  \includegraphics[width=\linewidth]{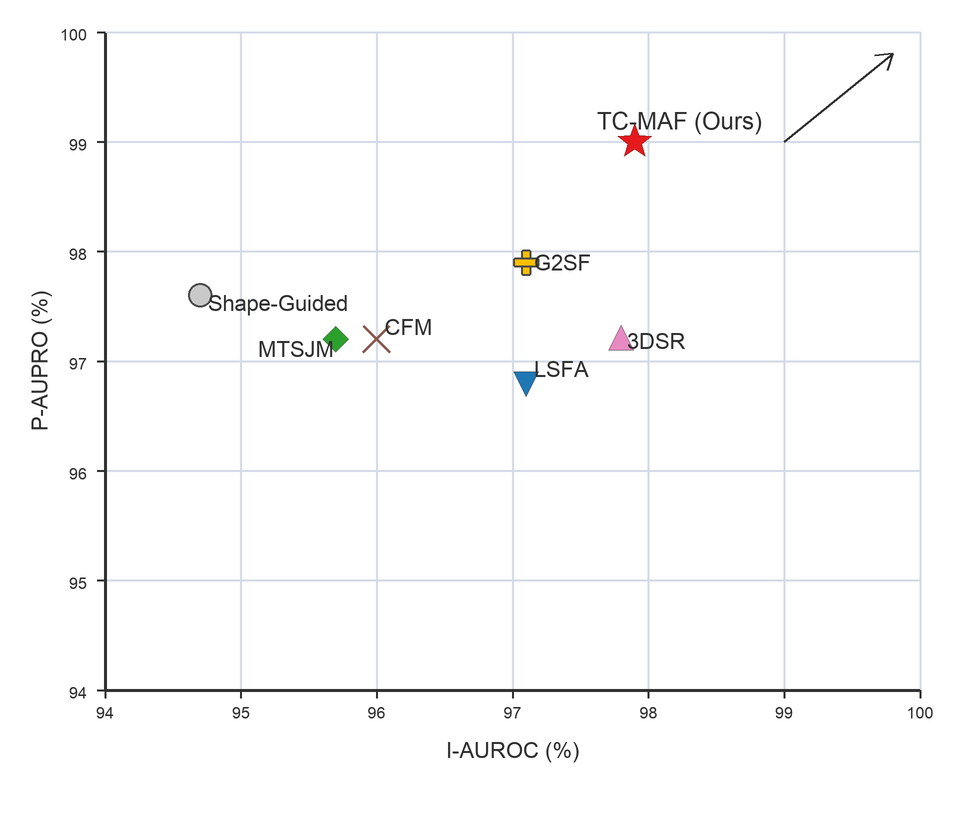}
  \caption{Detection-localization operating points on MVTec-3D. Each point shows one multimodal method in the I-AUROC--P-AUPRO plane. TC-MAF occupies the upper-right frontier among the compared methods, with the highest mean I-AUROC and P-AUPRO.}
  \Description{Scatter plot of MVTec-3D methods with I-AUROC on the horizontal axis and P-AUPRO on the vertical axis. TC-MAF appears as a red star at the upper-right frontier of the compared methods.}
  \label{fig:mvtec-operating-point}
\end{figure}

\section{Introduction}

Combining RGB appearance and 3D geometry improves industrial anomaly detection because many defects manifest more clearly in one modality than the other~\cite{wang2023m3dm,zavrtanik2024cheating,tao2025g2sf,ali2025mafr}. Once multiple evidence sources are available, however, the integration rule becomes important: auxiliary evidence may help substantially on some product categories and much less on others, while anomalous validation data for category-wise policy selection is usually unavailable. Prior methods advance representation learning, anomaly generation, or cross-modal interaction~\cite{wang2023m3dm,costanzino2024cfmad,sui2026cmdiad,long2026iumad,li2025find}, but usually aggregate branch outputs with fixed rules that are only lightly analyzed.

This issue becomes especially visible when a multimodal detector is paired with an RGB reconstruction branch. Reconstruction can recover subtle appearance anomalies that geometry may miss, yet its reliability is category-dependent: identity shortcuts emerge under diverse normal patterns~\cite{you2022uniad}, and stable multi-class reconstruction requires careful design~\cite{zavrtanik2021draem,guo2025dinomaly}. The resulting problem is not simply to add more evidence, but to incorporate that evidence without allowing it to overwrite the stronger multimodal base indiscriminately.

We address this with \textbf{TC-MAF}, a base-anchored multi-evidence fusion design. TC-MAF keeps the multimodal detector as the primary source, adds complementary Dinomaly evidence~\cite{guo2025dinomaly} through a constrained pixel-level mixture, and uses a small cross-modal consistency (CMC) term to reinforce regions where both modalities appear inconsistent. A lightweight training-dispersion confidence (TDC) term, computed from normal training statistics alone, scales the auxiliary contribution while keeping the fusion operator and its hyperparameters shared across categories.

Under this design, TC-MAF achieves \textbf{0.979} I-AUROC and \textbf{0.990} P-AUPRO on MVTec-3D, giving the best mean benchmark results among the compared methods. Systematic ablations show that the fusion structure outperforms several natural alternatives, while TDC contributes a smaller but repeatable gain over no calibration and arbitrary calibration. Table~\ref{tab:generalization} further shows that the same design remains effective when the auxiliary branch is replaced with ResAD~\cite{yao2024resad} or the base backbone with ResNet-18~\cite{he2016deep}. Additional results show that the method also transfers well to Eyecandies, few-shot settings, and a missing-3D setting.

\begin{figure*}[t]
  \centering
  \includegraphics[width=2\columnwidth]{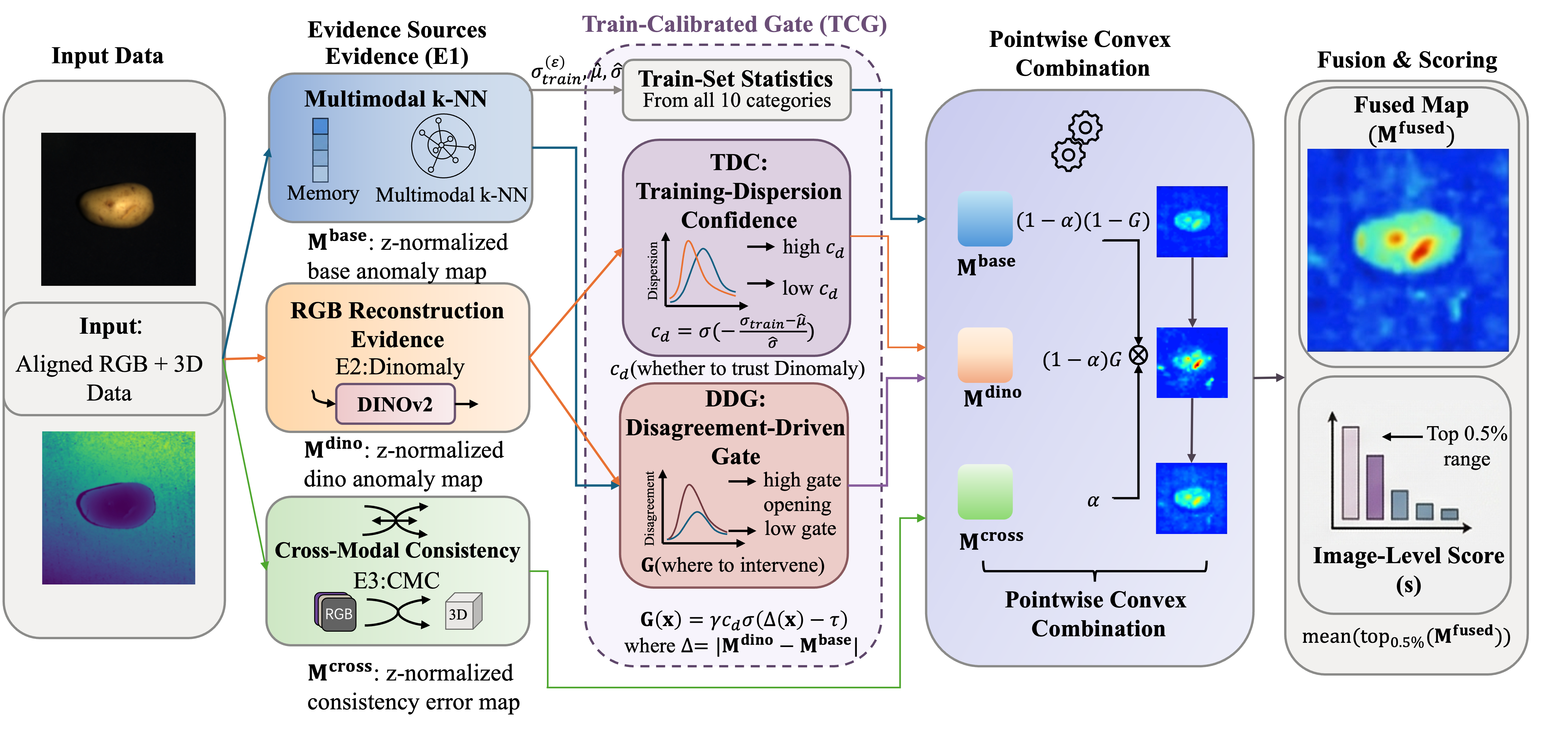}
  \caption{Schematic overview of TC-MAF. The multimodal base branch provides the anchor evidence, Dinomaly supplies complementary RGB reconstruction evidence, CMC adds a small cross-modal reinforcement cue, and TDC together with disagreement gating determines how strongly auxiliary evidence participates before the final image-level scoring step.}
  \Description{Schematic overview of the TC-MAF pipeline showing the multimodal base evidence, Dinomaly evidence, a CMC cue, Training-Dispersion Confidence, Disagreement-Driven Gate, and the final image-level scoring step.}
  \label{fig:method-overview}
\end{figure*}

Our contributions are as follows:
\begin{itemize}
  \item We propose a base-anchored multi-evidence fusion design for multimodal industrial AD, where auxiliary RGB reconstruction evidence is incorporated under one shared pixel-level fusion formula rather than through separate class-wise routing policies.
  \item We instantiate this design in TC-MAF: the multimodal detector remains the anchor, Dinomaly enters through a constrained mixture modulated by TDC and local disagreement, and CMC provides lightweight cross-modal reinforcement.
  \item We conduct systematic ablations over evidence sources, fusion alternatives, and calibration variants, show that a pooled-statistics variant remains close to the default setting, and further evaluate the same design under auxiliary-branch substitution (ResAD), backbone substitution (ResNet-18), cross-dataset evaluation (Eyecandies), few-shot settings, and a missing-3D setting.
\end{itemize}

\section{Related Work}

\textbf{Exploiting multimodality without cross-modal interference.}
Multimodal industrial AD methods agree that RGB and 3D cues are complementary but differ in where fusion is imposed. Benchmarks such as MVTec-3D and Eyecandies established the aligned RGB+3D setting for this problem~\cite{bergmann2022mvtec3d,bonfiglioli2022eyecandies}. BTF combines classical 3D descriptors with RGB features~\cite{horwitz2023btf}. CPMF studies point-cloud anomaly detection with complementary pseudo-multimodal features~\cite{cao2023cpmf}. M3DM couples patch-wise hybrid feature fusion with decision-layer fusion through multiple memory banks~\cite{wang2023m3dm}. 3DSR enhances 3D surface AD through depth simulation and a shared latent space~\cite{zavrtanik2024cheating}. CMDIAD addresses incomplete-modality inference via cross-modal distillation~\cite{sui2026cmdiad}. CFMAD learns cross-modal feature mappings for lightweight multimodal detection~\cite{costanzino2024cfmad}. G2SF explicitly studies geometry-guided score fusion~\cite{tao2025g2sf}. MTSJM combines mask training with teacher-student joint memory for multimodal industrial anomaly detection~\cite{liu2025multimodal}. 3D-ADNAS revisits multimodal fusion from an architectural perspective~\cite{long2025revisiting}. Collectively, these works broaden how multimodal evidence is extracted and fused, but the fusion rule itself is still typically fixed \emph{a priori} at the feature, architecture, or score level. This leaves limited room to modulate, at test time, \emph{how much} and \emph{where} each auxiliary branch should influence a particular sample and spatial region, even though fusion topology itself has recently been recognized as an important factor in 3D-AD~\cite{long2025revisiting}.

\textbf{Accuracy versus efficiency and deployment cost.}
Strong AD often involves a systems trade-off among accuracy, memory, and runtime. PaDiM and PatchCore achieve high accuracy through nominal modeling or retrieval, but at the cost of large feature stores~\cite{defard2021padim,roth2022patchcore}. SimpleNet shows that lightweight synthetic-anomaly discriminators can be competitive~\cite{liu2023simplenet}. AST introduces asymmetric student-teacher networks for efficient AD~\cite{rudolph2023ast}. PNI leverages position and neighborhood information for efficient detection~\cite{bae2023pni}. Yet computational efficiency alone does not address reliability mismatch across categories: under one shared fusion rule without per-class tuning, how should a system rebalance evidences whose utility varies from one category to another?

\textbf{Complementary reconstruction evidence versus shortcut risk.}
Reconstruction-style AD exposes fine-grained appearance deviations. DRAEM combines reconstruction and discrimination~\cite{zavrtanik2021draem}; RD4AD uses reverse distillation from one-class embedding~\cite{deng2022rd4ad}; CFLOW-AD employs conditional normalizing flows~\cite{gudovskiy2022cflow}; and TransFusion introduces a transparency-based diffusion model for anomaly detection~\cite{fucka2024transfusion}. UniAD identifies the identical-shortcut failure of reconstruction models, and Dinomaly shows that Transformer-based reconstruction can become a strong multi-class baseline~\cite{you2022uniad,guo2025dinomaly}. These works strengthen complementary appearance modeling, but when such evidence is paired with a multimodal detector, its reliability can still vary substantially across categories. A remaining practical question is therefore how to inject reconstruction-oriented auxiliary cues under a shared fusion rule without relying on test-set feedback or per-class tuning.

\textbf{Position of TC-MAF.}
The three threads above highlight a common need: once multiple evidences are available, the integration rule itself becomes a first-class design choice. TC-MAF addresses this by studying a base-anchored fusion rule with a simple training-only calibration term, and by validating that design against systematic fusion and calibration alternatives (Section~4).

\section{Method}

\subsection{Problem Setup and Overview}

We study unsupervised multimodal anomaly detection on MVTec-3D (10 categories, aligned RGB + 3D, normal-only real-data training). Following the standard benchmark protocol, inference is \emph{known-category}: the product category of a test sample is available, and the method outputs both an image-level score and a pixel-level anomaly map for that category.

A strong multimodal detector already captures much geometric and appearance evidence, yet it can under-react to subtle texture anomalies. A complementary RGB reconstruction branch can recover such cases, but uniform fusion is unreliable when that branch is unstable on certain categories. This leads to a more precise objective than simply estimating branch quality: under one shared fusion formula, incorporate auxiliary evidence without letting it dominate the stronger multimodal base.

TC-MAF realizes this objective through a base-anchored pixel-level fusion rule. A training-derived calibration term scales how strongly complementary evidence may participate, a disagreement-dependent gate determines where that participation is locally released, and a small CMC term reinforces regions with inter-modal inconsistency. All quantities derive from normal training statistics, while the fusion operator and its hyperparameters are shared across categories. Figure~\ref{fig:method-overview} summarizes the pipeline. Algorithm~\ref{alg:tcmaf} gives the complete procedure.

\paragraph{Notation.}
Let $\Omega$ denote the spatial grid and let $\mathbf{x} \in \Omega$ index a location. We write the three evidence maps as $\mathcal{E}=\{M^{\mathrm{base}}, M^{\mathrm{dino}}, M^{\mathrm{cross}}\}$ with $M^{(\cdot)} \in \mathbb{R}^{H \times W}$, and denote the deployed fusion operator by $\mathcal{F}(\mathcal{E}; c_d): \mathbb{R}^{H \times W \times 3} \to \mathbb{R}^{H \times W}$. By ``shared fusion formula,'' we mean that the functional form of $\mathcal{F}$ and its hyperparameters are fixed across categories under the known-category benchmark protocol. The default implementation uses category-specific normal statistics for branch normalization and TDC scaling, while a pooled-statistics variant is evaluated separately in Section~\ref{sec:ablation}.

\begin{figure*}[t]
  \centering
  \includegraphics[width=2\columnwidth]{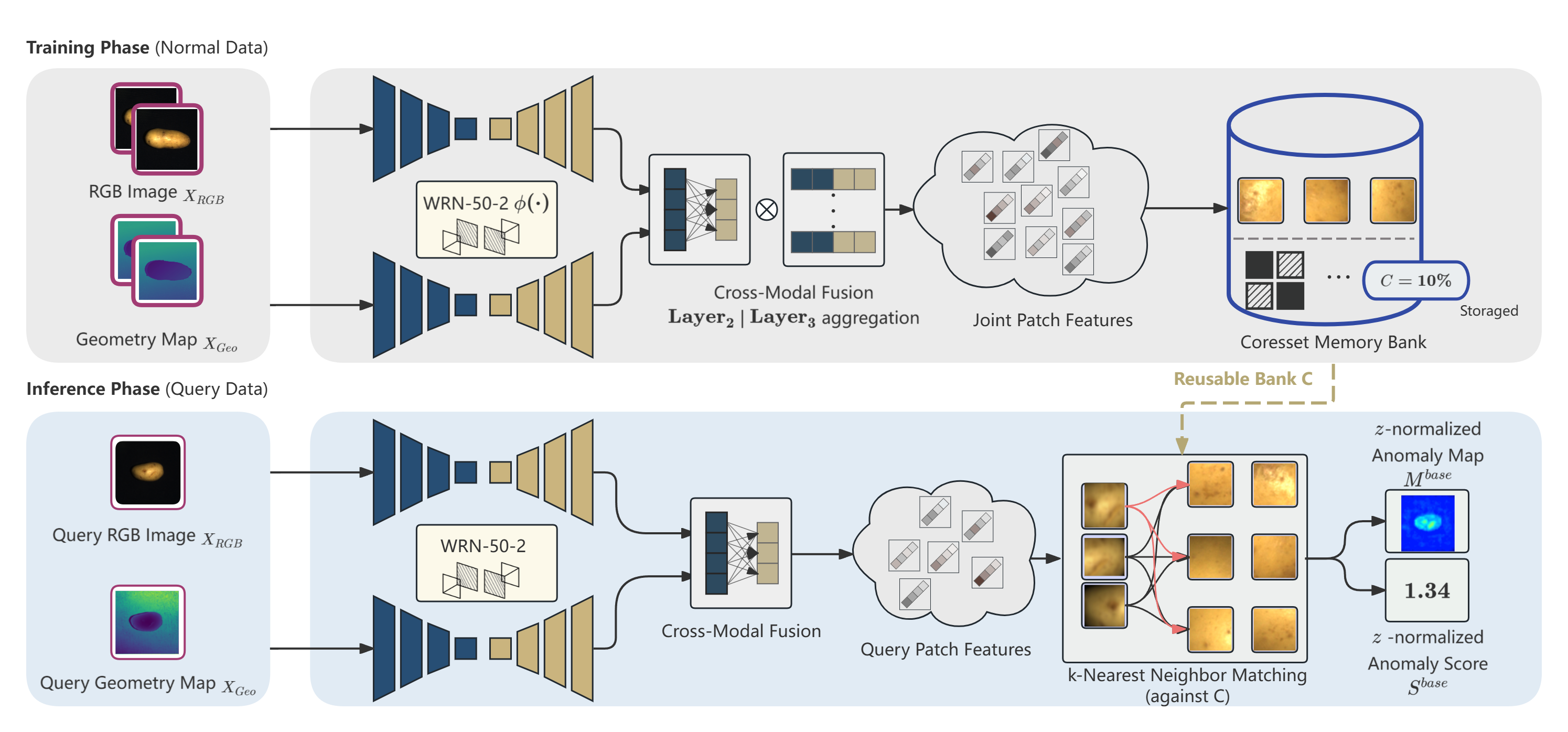}
  \caption{Refined schematic of the multimodal base branch used in TC-MAF. Training uses normal RGB and geometry pairs to build a coreset memory bank from fused joint patch features, while inference reuses that bank to score query patches through $k$-NN and produce $M^{\mathrm{base}}$ and $s^{\mathrm{base}}$.}
  \Description{A two-panel schematic of the multimodal base branch. The upper training panel shows normal RGB and geometry inputs passing through dual WRN-50-2 feature extraction and cross-modal fusion to form joint patch features, from which a 10 percent coreset memory bank is stored. The lower inference panel shows query RGB and geometry inputs passing through the same encoder, then being compared against the reused memory bank by k-nearest neighbors to produce the z-normalized base anomaly map and image-level score.}
  \label{fig:multimodal-base-branch}
\end{figure*}

\subsection{Evidence Sources}

TC-MAF combines three complementary evidence sources. The multimodal base and Dinomaly branches each produce a z-normalized anomaly map $M \in \mathbb{R}^{H \times W}$ and an image-level score $s \in \mathbb{R}$, while CMC provides an auxiliary normalized pixel map.

\paragraph{E1: Multimodal base evidence.}
The primary branch is a multimodal memory-bank-based detector that follows the Multimodal Memory k-NN design while using the implementation detailed in Appendix~A. It outputs a z-normalized base anomaly map $M^{\mathrm{base}}$ and score $s^{\mathrm{base}}$, serving as the default decision backbone. Figure~\ref{fig:multimodal-base-branch} shows a simplified view of this branch. Specifically, normal RGB--geometry pairs are encoded into joint patch features, a 10\% coreset memory bank is retained as the reference set, and query patches are scored against this bank by $k$-NN to produce the deployed base evidence.

\paragraph{E2: RGB reconstruction evidence.}
The auxiliary branch is Dinomaly~\cite{guo2025dinomaly}, a pure-RGB reconstruction model built on DINOv2~\cite{oquab2024dinov2} Transformer blocks. It produces a z-normalized anomaly map $M^{\mathrm{dino}}$ and score $s^{\mathrm{dino}}$. Dinomaly recovers subtle appearance anomalies that a geometry-heavy detector may miss, but its effectiveness varies across categories. TC-MAF therefore scales its contribution rather than treating it as a uniformly reliable peer.

\paragraph{E3: Cross-Modal Consistency (CMC)}
Let $\mathbf{r}(\mathbf{x})$ and $\mathbf{g}(\mathbf{x})$ denote the $\ell_2$-normalized RGB and geometry patch features at location $\mathbf{x}$, and let $f_{r\rightarrow g}, f_{g\rightarrow r}$ be the two train-set-fitted cross-modal mappers. TC-MAF measures bidirectional reconstruction inconsistency as
\begin{equation}\label{eq:cmc-def}
  e^{\mathrm{cross}}(\mathbf{x}) = \tfrac{1}{2}\!\left[\bigl(1 - \cos(\bar{\mathbf{g}}(\mathbf{x}), \mathbf{g}(\mathbf{x}))\bigr) + \bigl(1 - \cos(\bar{\mathbf{r}}(\mathbf{x}), \mathbf{r}(\mathbf{x}))\bigr)\right],
\end{equation}
where $\bar{\mathbf{g}}(\mathbf{x}) = \operatorname{norm}(f_{r\rightarrow g}(\mathbf{r}(\mathbf{x})))$ and $\bar{\mathbf{r}}(\mathbf{x}) = \operatorname{norm}(f_{g\rightarrow r}(\mathbf{g}(\mathbf{x})))$. The CMC map is the robustly normalized error map $M^{\mathrm{cross}}(\mathbf{x}) = (e^{\mathrm{cross}}(\mathbf{x}) - \tilde{\mu}_{\mathrm{cross}})/\tilde{\sigma}_{\mathrm{cross}}$, where $\tilde{\mu}_{\mathrm{cross}}$ and $\tilde{\sigma}_{\mathrm{cross}}$ are the training-set median and scaled MAD. Its role is purely auxiliary: it reinforces regions where both modalities exhibit unusual cross-modal inconsistency.

\subsection{Adaptive Fusion with a Shared Fusion Rule}

Our deployed fusion rule keeps the multimodal base branch as the anchor, uses a lightweight training-derived calibration term to scale Dinomaly's contribution, and uses local disagreement to decide where that contribution should be spatially released. The construction keeps the base branch dominant while still allowing complementary evidence to correct it where needed.

\paragraph{Training-Dispersion Confidence (TDC)}
In the normal-only industrial setting, auxiliary usefulness cannot be calibrated from anomalous validation data. We therefore use a training-derived calibration term based on the dispersion of Dinomaly's image-level scores on normal training samples. This term acts as a coarse prior for scaling the auxiliary branch inside the fusion rule rather than as a direct predictor of downstream gain. In the default known-category setting, each category $c$ has its own training-score dispersion $\sigma_{\mathrm{train}}^{(c)}$, computed from Dinomaly's image-level scores on the normal training samples of that category, and converted into a bounded scalar:
\begin{equation}\label{eq:dino-conf}
  c_d^{(c)} = \sigma\!\left(-\frac{\sigma_{\mathrm{train}}^{(c)} - \hat{\mu}}{\hat{\sigma}}\right),
\end{equation}
where $\sigma(\cdot)$ is the sigmoid function, and $\hat{\mu}$, $\hat{\sigma}$ are the median and standard deviation of $\{\sigma_{\mathrm{train}}^{(c)}\}_{c=1}^{C}$ computed across the training classes. The two summary scalars $\hat{\mu}$ and $\hat{\sigma}$ are fixed constants shared across deployment. Since $\sigma(\cdot)$ is monotonically increasing, $c_d^{(c)}$ is strictly decreasing in training dispersion, with midpoint $c_d^{(c)}=0.5$ at $\sigma_{\mathrm{train}}^{(c)}=\hat{\mu}$. In the equations below we write $c_d$ for the category-specific value applied at inference. In practice, TDC is kept because it is simple, training-only, and improves over fixed or random calibration by about 0.85pp I-AUROC (Section~\ref{sec:ablation}). A pooled-statistics variant, which replaces the default per-class normalization statistics and category-specific $\sigma_{\mathrm{train}}^{(c)}$ with pooled training statistics, is evaluated separately and remains close to the default setting in our ablation.

\paragraph{Locally gated evidence aggregation.}
At each location $\mathbf{x}$, TC-MAF combines the training-derived scaling term with spatial disagreement. Let $\Delta(\mathbf{x}) = \lvert M^{\mathrm{dino}}(\mathbf{x}) - M^{\mathrm{base}}(\mathbf{x})\rvert$ and let $\tau$ denote the disagreement threshold (implemented as $\tau=1$). We use
\begin{equation}\label{eq:gate-map}
  G(\mathbf{x}) = \gamma \, c_d \, \sigma\!\bigl(\Delta(\mathbf{x}) - \tau\bigr),
\end{equation}
as the deployed gate map. Since $\sigma(\cdot)$ is monotonically increasing, the disagreement term is monotonically increasing in disagreement magnitude. We use $G(\mathbf{x})$ as a spatially varying mixing coefficient that releases more auxiliary correction where the Dinomaly and base maps disagree more strongly, while still being scaled by $c_d$.

The fused map is then written in one step as
\begin{equation}
\label{eq:fused-convex}
\begin{aligned}
M^{\mathrm{fused}}(\mathbf{x}) =\;&
\underbrace{(1-\alpha)(1-G(\mathbf{x}))}_{w^{\mathrm{base}}(\mathbf{x})}\,
M^{\mathrm{base}}(\mathbf{x}) \\
&+ \underbrace{(1-\alpha)G(\mathbf{x})}_{w^{\mathrm{dino}}(\mathbf{x})}\,
M^{\mathrm{dino}}(\mathbf{x})
+ \underbrace{\alpha}_{w^{\mathrm{cross}}}\,
M^{\mathrm{cross}}(\mathbf{x}).
\end{aligned}
\end{equation}
Because $c_d \in (0,1)$, $\sigma(\Delta(\mathbf{x})-\tau) \in (0,1)$, and $\gamma<1$, we have $0 \leq G(\mathbf{x}) < 1$; therefore all three coefficients are nonnegative and sum to one, so Eq.~\ref{eq:fused-convex} is a pointwise convex combination of the three evidence sources. The base detector always remains active, and auxiliary correction is explicitly bounded by the gate.

\paragraph{Design rationale.}
TDC provides a training-derived scaling factor, disagreement provides a spatial release signal, and the convex mixture keeps the base detector as the primary source. Together, these components allow complementary evidence to contribute where useful while preventing it from overwriting the base indiscriminately. Section~\ref{sec:ablation} validates this design: several natural alternatives---static averaging, score-level fusion, rank-level fusion, and a learned gate---all fall short. Among calibration variants, constant or random $c_d$ drops I-AUROC by ${\sim}0.85$pp, while the pooled-statistics variant remains close to Full.

\subsection{Image-Level Scoring}

The image-level score is computed from the fused map as
\begin{equation}\label{eq:topk-score}
  s = \mathrm{mean}\!\bigl(\mathrm{top}_{0.5\%}(M^{\mathrm{fused}})\bigr).
\end{equation}
Equivalently, if $P=HW$, $m=\max\!\bigl(1, \lfloor 0.005P \rfloor\bigr)$, and $M^{\mathrm{fused}}_{(1)} \geq \cdots \geq M^{\mathrm{fused}}_{(P)}$ are the sorted pixel values, then
\begin{equation}\label{eq:topk-order}
  s = \frac{1}{m}\sum_{t=1}^{m} M^{\mathrm{fused}}_{(t)}.
\end{equation}
This image-level scoring step affects only I-AUROC ranking; it does not modify pixel maps, retrain any branch, or change P-AUPRO. The same fused map $M^{\mathrm{fused}}$ is used directly for pixel-level localization. We use a fixed top-0.5\% readout throughout the paper to convert fused maps into image-level scores. This readout is part of the deployed evaluation protocol for TC-MAF and is kept unchanged across all experiments.

\begin{algorithm}[t]
\caption{TC-MAF: Adaptive Fusion with a Shared Fusion Rule}\label{alg:tcmaf}
\small
\begin{algorithmic}[1]
\Statex \textbf{Training phase} (normal data only):
\State Dinomaly (pre-trained); for each class $c$, compute score std $\sigma_{\mathrm{train}}^{(c)}$ on normal training samples
\State Compute shared $\hat{\mu}, \hat{\sigma}$ $\leftarrow$ median, std of $\{\sigma_{\mathrm{train}}^{(c)}\}_{c=1}^{C}$ \Comment{or use pooled training statistics}
\State Build multimodal base memory bank; fit CMC mappers
\State Store branch normalization statistics \Comment{mean/std for base and Dinomaly; median/MAD for CMC}
\Statex
\Statex \textbf{Inference phase} (shared fusion formula, all classes):
\State $M^{\mathrm{base}}, M^{\mathrm{dino}}, M^{\mathrm{cross}}$ $\leftarrow$ branch-normalized maps \Comment{z-normalized for base/Dinomaly, robustly normalized for CMC}
\State $c_d \leftarrow \sigma\!\bigl(-(\sigma_{\mathrm{train}}^{(c)} - \hat{\mu})/\hat{\sigma}\bigr)$ \Comment{TDC, Eq.~\ref{eq:dino-conf}}
\State $\Delta \leftarrow \lvert M^{\mathrm{dino}} - M^{\mathrm{base}} \rvert$ \Comment{elementwise disagreement}
\State $G \leftarrow \gamma \, c_d \, \sigma(\Delta - \tau)$ \Comment{vectorized gate map, $\mathcal{O}(HW)$, Eq.~\ref{eq:gate-map}}
\State $M^{\mathrm{fused}} \leftarrow (1{-}\alpha)(1{-}G) \odot M^{\mathrm{base}} + (1{-}\alpha)G \odot M^{\mathrm{dino}} + \alpha M^{\mathrm{cross}}$ \Comment{vectorized fusion, $\mathcal{O}(HW)$, Eq.~\ref{eq:fused-convex}}
\State $s \leftarrow \mathrm{mean}\!\bigl(\mathrm{top}_{0.5\%}(M^{\mathrm{fused}})\bigr)$ \Comment{Eq.~\ref{eq:topk-score}}
\State \Return $M^{\mathrm{fused}},\; s$
\end{algorithmic}
\end{algorithm}

\begin{figure}[!htbp]
  \centering
  \includegraphics[width=\columnwidth]{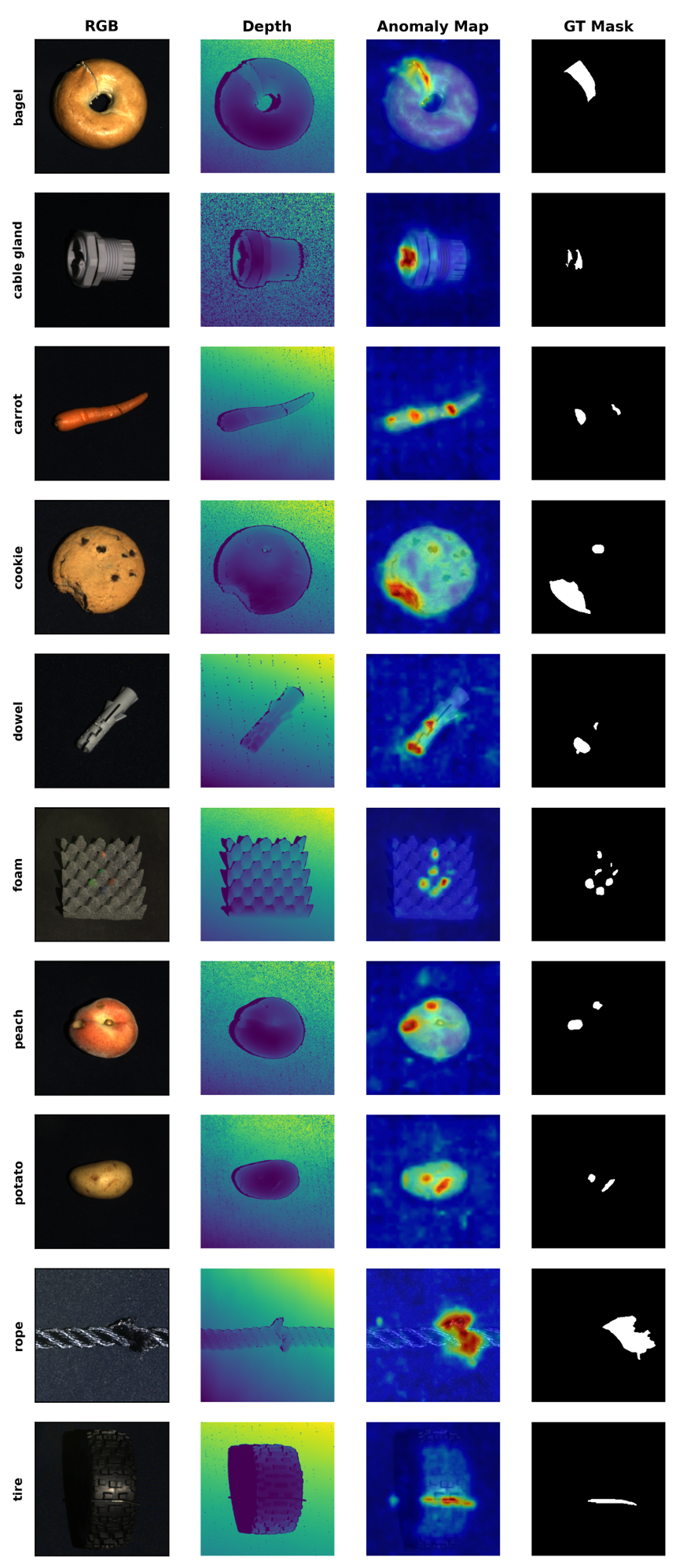}
  \caption{Full-class qualitative overview on MVTec-3D. Each row shows one representative anomalous sample for one class with RGB, depth, the final TC-MAF anomaly map, and the ground-truth mask.}
  \Description{Full-class qualitative overview on MVTec-3D, with one row per class and four columns showing RGB, depth, TC-MAF anomaly map, and ground-truth mask.}
  \label{fig:mvtec3d-fullclass}
\end{figure}

\section{Experiments}
\begin{table*}[!htbp]
  \centering
  \caption{Expanded class-wise comparison on MVTec-3D across 3D, RGB, and 3D+RGB settings. Each cell reports I-AUROC/P-AUPRO.}
  \label{tab:classwise-comparison}
  \small
  \setlength{\tabcolsep}{2pt}
  \renewcommand{\arraystretch}{1.02}
  \begin{tabular}{@{}llccccccccccc@{}}
    \toprule
    Setting & Method & Bagel & C.Gland & Carrot & Cookie & Dowel & Foam & Peach & Potato & Rope & Tire & Mean \\
    \midrule
    \multirow{5}{*}{3D}
    & Shape-G.'23~\cite{chu2023shapeguided} & .983/\best{.974} & .682/.871 & \best{.978}/\second{.981} & \best{.998}/\second{.924} & \best{.960}/.898 & .737/\second{.773} & \best{.993}/\second{.978} & \best{.979}/\second{.983} & .966/.955 & .871/\second{.969} & .916/.931 \\
    & 3DSR'24~\cite{zavrtanik2024cheating} & .945/.922 & \second{.835}/.872 & .969/\best{.984} & .857/.859 & \second{.955}/\second{.940} & \second{.880}/.714 & .963/.970 & .934/.978 & \best{.998}/\best{.977} & .888/.858 & \second{.922}/.907 \\
    & LSFA'24~\cite{tu2024lsfa} & \best{.986}/\best{.974} & .669/\second{.887} & .973/\second{.981} & .990/.921 & .950/.901 & .802/\second{.773} & .961/\best{.982} & \second{.964}/\second{.983} & \second{.967}/.959 & \best{.944}/\best{.981} & .921/\second{.934} \\
    & MTSJM'25~\cite{liu2025multimodal} & \second{.985}/.929 & .728/.855 & .961/.971 & \second{.991}/.887 & .876/.878 & .773/.764 & .906/.940 & .935/.971 & .932/.937 & .856/.951 & .894/.908 \\
    & TC-MAF (Ours) & .972/\second{.963} & \best{.913}/\best{.942} & \second{.974}/.975 & .881/\best{.936} & .934/\best{.984} & \best{.907}/\best{.887} & \second{.983}/.972 & .949/\best{.985} & \second{.967}/\second{.969} & \second{.898}/.903 & \best{.938}/\best{.952} \\
    \midrule
    \multirow{5}{*}{RGB}
    & Shape-G.'23~\cite{chu2023shapeguided} & .911/.946 & \second{.936}/.972 & .883/.960 & .662/\best{.914} & \second{.974}/\second{.958} & .772/.776 & .785/.937 & .641/.949 & .884/.956 & .706/.957 & .815/.933 \\
    & 3DSR'24~\cite{zavrtanik2024cheating} & .844/.923 & .930/.970 & \second{.964}/\second{.979} & \second{.794}/.859 & \best{.998}/\best{.979} & \best{.904}/\second{.894} & .938/.943 & \second{.730}/.951 & \best{.978}/.964 & \best{.900}/\best{.980} & \second{.898}/.944 \\
    & LSFA'24~\cite{tu2024lsfa} & .951/.957 & .920/\second{.976} & .911/.970 & .762/.912 & .961/.934 & .770/.851 & .930/.960 & .675/.957 & \second{.938}/\second{.970} & .787/.961 & .861/.945 \\
    & MTSJM'25~\cite{liu2025multimodal} & \second{.971}/\second{.976} & .908/.969 & .942/.975 & .731/\best{.914} & .967/.949 & .795/.869 & \best{.984}/\second{.975} & .608/\second{.963} & .838/\second{.970} & .823/\second{.966} & .858/\second{.953} \\
    & TC-MAF (Ours) & \best{.973}/\best{.982} & \best{.967}/\best{.977} & \best{.979}/\best{.987} & \best{.861}/\second{.913} & .972/.943 & \second{.851}/\best{.901} & \second{.962}/\best{.980} & \best{.934}/\best{.976} & .936/\best{.981} & \second{.895}/.946 & \best{.933}/\best{.959} \\
    \midrule
    \multirow{7}{*}{3D+RGB}
    & Shape-G.'23~\cite{chu2023shapeguided} &
    .986/.981 & .894/.973 & .983/.982 & .991/.971 & .976/.962 & .857/\second{.978} & \second{.990}/.981 & .965/.983 & .960/.974 & .869/.975 & .947/.976 \\
    & 3DSR'24~\cite{zavrtanik2024cheating} &
    .981/.964 & .867/.966 & \best{.996}/.981 & .981/.942 & \best{1.000}/\second{.980} & \best{.994}/.973 & .986/.981 & .978/.977 & \best{1.000}/\best{.979} & \best{.995}/.979 & \second{.978}/.972 \\
    & CFM'24~\cite{costanzino2024cfmad} &
    .988/.980 & .875/.966 & .984/.982 & \second{.992}/.947 & \second{.997}/.959 & .924/.967 & .964/.982 & .949/.983 & .979/.976 & .950/.982 & .960/.972 \\
    & LSFA'24~\cite{tu2024lsfa} &
    \best{1.000}/\second{.986} & \second{.939}/.974 & .982/.981 & .989/.946 & .961/.925 & .951/.941 & .983/\second{.983} & .962/.983 & \second{.989}/.974 & .951/\second{.983} & .971/.968 \\
    & G2SF'25~\cite{tao2025g2sf} &
    \second{.997}/.982 & .923/.977 & \second{.993}/.982 & .967/\second{.979} & .966/.971 & \second{.991}/.976 & \best{.994}/.982 & \best{.988}/.983 & .966/\second{.978} & .922/.981 & .971/\second{.979} \\
    & MTSJM'25~\cite{liu2025multimodal} &
    \best{1.000}/.984 & .931/\second{.981} & .985/\second{.983} & \best{.994}/.968 & .968/.939 & .899/.951 & .986/.977 & .947/\second{.985} & .962/.974 & .897/.973 & .957/.972 \\
    \midrule
    & TC-MAF (Ours) &
    .992/\best{.993} & \best{.998}/\best{.997} & .992/\best{.997} & .953/\best{.989} & .992/\best{.991} & .952/\best{.979} & .988/\best{.998} & \second{.979}/\best{.998} & .984/.963 & \second{.958}/\best{.992} & \best{.979}/\best{.990} \\
    \bottomrule
  \end{tabular}
\end{table*}

\begin{table}[t]
  \centering
  \caption{Comparison with selected 3D+RGB multimodal methods on MVTec-3D.}
  \label{tab:main-comparison}
  \begin{tabular}{lcc}
    \toprule
    Method & I-AUROC & P-AUPRO \\
    \midrule
    Shape-Guided'23~\cite{chu2023shapeguided} & 0.947 & 0.976 \\
    3DSR'24~\cite{zavrtanik2024cheating} & \second{0.978} & 0.972 \\
    CFM'24~\cite{costanzino2024cfmad} & 0.960 & 0.972 \\
    LSFA'24~\cite{tu2024lsfa} & 0.971 & 0.968 \\
    G2SF'25~\cite{tao2025g2sf} & 0.971 & \second{0.979} \\
    MTSJM'25~\cite{liu2025multimodal} & 0.957 & 0.972 \\
    \midrule
    TC-MAF (Ours) & \best{0.979} & \best{0.990} \\
    \bottomrule
  \end{tabular}
\end{table}
\subsection{Experimental Setup}

\paragraph{Dataset.}
We evaluate on MVTec-3D~\cite{bergmann2022mvtec3d}, which contains 10 industrial product categories with aligned RGB images and organized 3D point clouds. Each category provides only normal samples for training; the test set includes both normal and anomalous samples with pixel-level ground-truth masks. As in the standard benchmark protocol, evaluation is performed category by category.

\paragraph{Metrics.}
We report image-level AUROC (I-AUROC) for detection, pixel-level AUROC (P-AUROC), and pixel-level AUPRO (P-AUPRO), following the standard MVTec-3D protocol. Image-level scores are computed with Eq.~\ref{eq:topk-score} throughout the paper.

\paragraph{Baselines.}
We compare against representative 3D+RGB multimodal baselines on MVTec-3D, including Shape-Guided~\cite{chu2023shapeguided}, 3DSR~\cite{zavrtanik2024cheating}, CFM~\cite{costanzino2024cfmad}, LSFA~\cite{tu2024lsfa}, G2SF~\cite{tao2025g2sf}, and MTSJM~\cite{liu2025multimodal}.

\paragraph{Implementation.}
TC-MAF uses the configuration described in Section~3: Dinomaly~\cite{guo2025dinomaly} with DINOv2~\cite{oquab2024dinov2} ViT-S/14 at $448{\times}448$, multimodal base branch with dual WRN-50-2~\cite{zagoruyko2016wide} at $288{\times}288$, maximum gate $\gamma{=}0.6$, and CMC weight $\alpha{=}0.15$. All 10 classes share identical fusion settings and training procedures; branch-specific normalization statistics are computed per class by default, while the pooled-statistics variant is evaluated separately in Table~\ref{tab:ablation-extended}.

\paragraph{Protocol scope and comparison notes.}
All fusion equations and hyperparameters are fixed from training-set statistics and applied identically across all 10 classes; no class-specific fusion formula or threshold optimization is performed. We follow the standard MVTec-3D metric definitions (same \texttt{roc\_auc\_score}, same PRO pipeline with 200 thresholds, FPR$<0.3$, morphological dilation). The pooled-statistics variant and the missing-3D setting are both summarized in Table~\ref{tab:ablation-extended}.

\subsection{Benchmark Positioning}

Table~\ref{tab:main-comparison} summarizes TC-MAF together with representative methods on MVTec-3D.

In Table~\ref{tab:main-comparison}, TC-MAF achieves the highest mean I-AUROC and P-AUPRO among the compared methods. The lead is 0.1pp on I-AUROC over 3DSR (0.979 vs.~0.978) and 1.1pp on P-AUPRO over G2SF (0.990 vs.~0.979). The main comparison focuses on I-AUROC and P-AUPRO; P-AUROC results are provided separately in Appendix~\ref*{app:pauroc}.

Class-wise, TC-MAF is especially strong on Cable Gland and maintains near-ceiling localization on most categories, with Rope as the main localization exception, while Cookie and Foam remain the harder categories in the class-wise I-AUROC breakdown. This pattern is consistent with the method's structure: TC-MAF benefits most when locally concentrated corrective evidence can complement the base, whereas diffuse anomaly concentration remains harder for image-level ranking. Detailed class-wise breakdown is presented in Table~\ref{tab:classwise-comparison}.

\subsection{Controlled Evidence: Component Ablation}\label{sec:ablation}

Table~\ref{tab:ablation-extended} presents the systematic ablation study for TC-MAF. All rows use the same evaluation protocol. The study is organized into four groups: (A)~evidence ablation, (B)~fusion-strategy alternatives, (C)~calibration variants, and (D)~extended settings. In Group~A, ``w/o TDC'' removes the training-derived scaling term from the deployed gate; in Group~C, the same gate form is retained but $c_d$ is replaced with non-informative fixed or randomized scalars.

\begin{table}[t]
  \centering
  \caption{Systematic ablation on MVTec-3D. All variants use the same evaluation protocol. Groups: (A)~evidence ablation, (B)~fusion-strategy alternatives, (C)~calibration variants, (D)~extended settings. $\Delta$I is the I-AUROC difference from Full (pp).}
  \label{tab:ablation-extended}
  \small
  \begin{tabular}{llccc}
    \toprule
    & Variant & I-AUROC & $\Delta$I & P-AUPRO \\
    \midrule
    & TC-MAF (Full) & \textbf{.9788} & --- & \textbf{.9901} \\
    \midrule
    \multirow{3}{*}{\rotatebox{90}{\scriptsize A}}
    & w/o Dinomaly (base only) & .9554 & $-$2.34 & .9807 \\
    & w/o CMC & .9644 & $-$1.44 & .9891 \\
    & w/o TDC & .9709 & $-$0.79 & .9881 \\
    \midrule
    \multirow{5}{*}{\rotatebox{90}{\scriptsize B}}
    & static pixel mean & .9488 & $-$3.00 & .9871 \\
    & score-level weighted avg & .9521 & $-$2.67 & .9875 \\
    & rank-level fusion & .9634 & $-$1.54 & .9877 \\
    & learned logistic gate & .9571 & $-$2.17 & .9872 \\
    & w/o disagreement gate & .9690 & $-$0.98 & .9878 \\
    \midrule
    \multirow{3}{*}{\rotatebox{90}{\scriptsize C}}
    & constant $c_d$ & .9702 & $-$0.86 & .9898 \\
    & randomized $c_d$ & .9704 & $-$0.84 & .9896 \\
    & pooled-statistics variant & .9782 & $-$0.06 & .9892 \\
    \midrule
    \multirow{1}{*}{\rotatebox{90}{\scriptsize D}}
    & missing 3D modality & .9330 & $-$4.58 & .9586 \\
    \bottomrule
  \end{tabular}
\end{table}

\begin{figure*}[!htbp]
  \centering
  \includegraphics[width=0.95\textwidth]{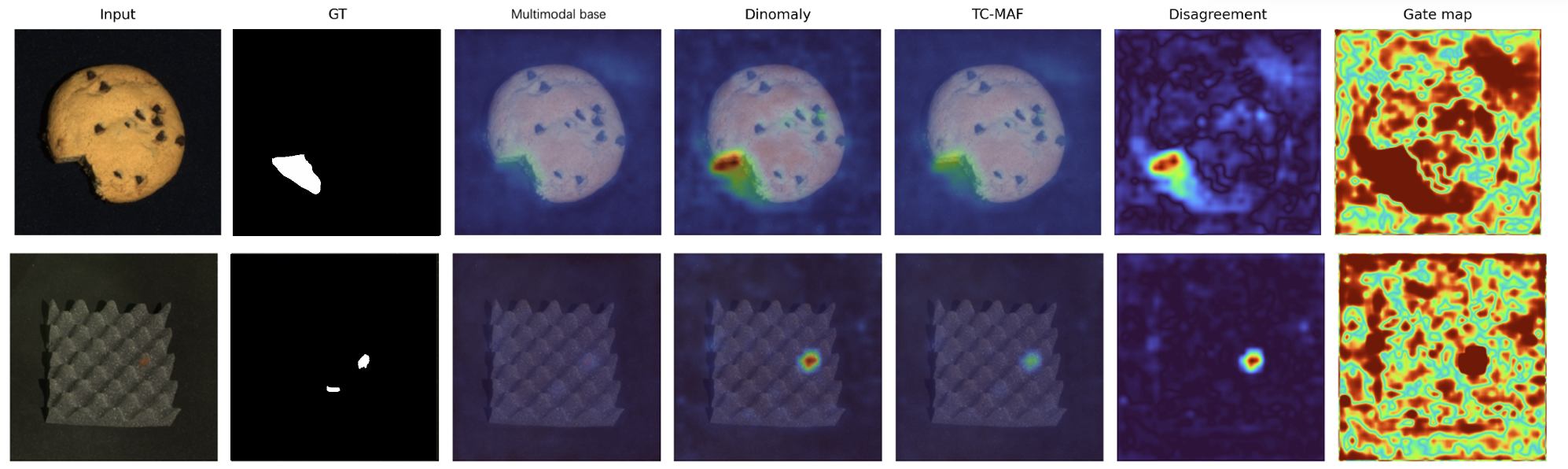}
  \caption{Weak-category visualization for Cookie and Foam. The examples provide a larger view of the branch-wise responses and the final TC-MAF fused maps, illustrating the diffuse anomaly evidence that makes image-level ranking difficult on these classes while still allowing strong pixel-level localization.}
  \Description{Weak-category visualization on Cookie and Foam, showing enlarged branch-wise comparisons and the final TC-MAF fused anomaly maps.}
  \label{fig:appendix-qualitative-weak}
\end{figure*}

The ablation reveals a clear hierarchy. \textbf{Group~A (evidence ablation):} Within the core fusion-component ablations, removing Dinomaly causes the largest drop ($-$2.34pp I-AUROC, $-$0.94pp P-AUPRO), indicating that complementary reconstruction evidence is the main additional source of gain. CMC adds 1.44pp to I-AUROC, while TDC contributes a smaller 0.79pp improvement. \textbf{Group~B (fusion-strategy alternatives):} All Group~B variants retain the same three evidence sources and TDC; only the way they are combined is changed. Static pixel mean drops I-AUROC by 3.00pp; a learned logistic gate drops it by 2.17pp; score-level weighted averaging ($-$2.67pp) and rank-level fusion ($-$1.54pp) also fall short of the deployed pixel-level rule. Removing the disagreement gate costs 0.98pp. These comparisons indicate that the main benefit comes from the fusion structure rather than from generic aggregation. \textbf{Group~C (calibration variants):} Constant $c_d$ ($-$0.86pp) and randomized $c_d$ ($-$0.84pp) both degrade I-AUROC, showing that the training-derived signal is more useful than an arbitrary scalar, even though its average contribution is modest. The pooled-statistics variant, which replaces the default per-class normalization statistics and category-specific $\sigma_{\mathrm{train}}^{(c)}$ with pooled training statistics, remains close to Full ($-$0.06pp I-AUROC, $-$0.09pp P-AUPRO), suggesting that the rule is not tightly tied to category-specific calibration statistics. \textbf{Group~D:} Under the missing-3D setting, I-AUROC decreases by 4.58pp and P-AUPRO by 3.15pp. Performance remains below the full RGB+3D setting, but the RGB-driven evidence and bounded fusion rule still retain useful detection and localization ability when 3D is unavailable.

\paragraph{Supporting analysis for the training-derived calibration term.}
We interpret TDC as a lightweight training-only calibrator rather than as the main driver of performance. In Table~\ref{tab:ablation-extended}, removing TDC reduces mean I-AUROC by 0.79pp, indicating a secondary but consistent contribution on top of the shared fusion rule. The complementary per-class analysis in Appendix~\ref*{app:tdc-correlation} supports the same mechanism: training dispersion correlates with the observed fusion gain over the base branch ($\rho{=}0.66$, $p{=}0.038$), while TDC confidence correlates with the magnitude of the TDC-specific effect ($\rho{=}0.68$, $p{=}0.029$). Together, these results suggest that TDC primarily regulates auxiliary participation under a fixed fusion rule instead of acting as a uniform additive boost. Alternative train-dispersion proxies are summarized in Table~\ref*{tab:proxy-comparison}.

\subsection{Additional Results}\label{sec:supporting}

\paragraph{Few-shot learning.}

To evaluate TC-MAF under limited training data, we conduct few-shot experiments on MVTec-3D with 5, 10, and 50 training samples per class. Table~\ref{tab:fewshot-benchmark} reports TC-MAF results under the same shot settings together with prior methods.

\begin{table}[t]
  \centering
  \caption{Few-shot/full-shot results on MVTec-3D. TC-MAF is evaluated under the same shot settings.}
  \label{tab:fewshot-benchmark}
  \small
  \resizebox{\columnwidth}{!}{
  \begin{tabular}{lcccccccc}
    \toprule
    & \multicolumn{4}{c}{I-AUROC} & \multicolumn{4}{c}{P-AUPRO} \\
    \cmidrule(lr){2-5} \cmidrule(lr){6-9}
    Method & 5-shot & 10-shot & 50-shot & Full & 5-shot & 10-shot & 50-shot & Full \\
    \midrule
    BTF'23~\cite{horwitz2023btf} & 0.671 & 0.695 & 0.806 & 0.865 & 0.920 & 0.928 & 0.947 & 0.959 \\
    AST'23~\cite{rudolph2023ast} & 0.680 & 0.689 & 0.794 & 0.937 & 0.903 & 0.835 & 0.929 & 0.944 \\
    M3DM'23~\cite{wang2023m3dm} & 0.822 & 0.845 & 0.907 & 0.945 & 0.937 & 0.943 & 0.955 & 0.964 \\
    CFM'24~\cite{costanzino2024cfmad} & 0.811 & 0.845 & 0.906 & 0.960 & \second{0.949} & \second{0.954} & \second{0.965} & \second{0.972} \\
    LSFA'24~\cite{tu2024lsfa} & \best{0.834} & \second{0.871} & \second{0.926} & \second{0.971} & 0.936 & 0.943 & 0.962 & 0.968 \\
    TC-MAF (Ours) & \second{0.830} & \best{0.873} & \best{0.941} & \best{0.979} & \best{0.952} & \best{0.962} & \best{0.973} & \best{0.990} \\
    \bottomrule
  \end{tabular}
  }
\end{table}

Table~\ref{tab:fewshot-benchmark} shows that TC-MAF delivers the best P-AUPRO across all shot settings and the best I-AUROC from 10-shot onward. At 5-shot, LSFA is marginally higher on I-AUROC (0.834 vs.~0.830), but TC-MAF still keeps the strongest localization result. Under reduced normal training data, the same fusion design therefore remains competitive in detection and consistently strong in localization.

Per-class few-shot results are provided in Appendix~\ref*{app:fewshot-classwise}.

\paragraph{Pooled-statistics variant.}

A natural question is whether the fusion rule is tightly coupled to category-specific calibration statistics. To test this, we replace the default per-class normalization statistics and category-specific $\sigma_{\mathrm{train}}^{(c)}$ with pooled training statistics computed across all 10 classes. Table~\ref{tab:ablation-extended} (Group~C, ``pooled-statistics variant'') shows that this variant remains close to Full performance ($-$0.06pp I-AUROC, $-$0.09pp P-AUPRO). This result suggests that, within the standard benchmark protocol, the rule is not tightly dependent on category-specific calibration statistics. The default per-class setup is retained for consistency with the known-category evaluation protocol rather than because it yields a clear advantage in this ablation.

\paragraph{Missing-3D setting.}

We further evaluate the case where the 3D input is unavailable at inference. Table~\ref{tab:ablation-extended} (Group~D) shows a clear reduction from Full performance ($-$4.58pp I-AUROC, $-$3.15pp P-AUPRO). The result indicates that geometry remains useful, while the RGB pathway still carries substantial anomaly information under the same shared fusion rule.

\paragraph{Generalization across branches and backbones.}

A key concern is whether TC-MAF's bounded fusion design is specific to the Dinomaly~\cite{guo2025dinomaly} + WRN-50-2~\cite{zagoruyko2016wide} combination or generalizes to other components. We test two substitutions while keeping the fusion rule identical: (i)~replacing Dinomaly with ResAD~\cite{yao2024resad}, a residual-based RGB-only detector with a different inductive bias, and (ii)~replacing the WRN-50-2 base backbone with ResNet-18~\cite{he2016deep}. For each substitution, we compare the full bounded fusion rule against static pixel mean and a learned gate under the same evaluation protocol.

\begin{table}[t]
  \centering
  \caption{Generalization of the bounded fusion rule across auxiliary branches and base backbones on MVTec-3D. Each block replaces one component while keeping the fusion rule identical. $\Delta$I is the I-AUROC gap from bounded fusion to the given baseline (pp).}
  \label{tab:generalization}
  \small
  \begin{tabular}{llccc}
    \toprule
    Setting & Fusion & I-AUROC & P-AUPRO & $\Delta$I \\
    \midrule
    \multicolumn{5}{l}{\emph{Default: Dinomaly + WRN-50-2 base}} \\
    & bounded (Full) & \textbf{.979} & \textbf{.990} & --- \\
    & static mean & .949 & .987 & $-$3.00 \\
    & learned gate & .957 & .987 & $-$2.17 \\
    \midrule
    \multicolumn{5}{l}{\emph{Alt.\ auxiliary: ResAD + WRN-50-2 base}} \\
    & bounded & \textbf{.946} & \textbf{.891} & --- \\
    & static mean & .873 & .864 & $-$7.30 \\
    & learned gate & .915 & .783 & $-$3.10 \\
    \midrule
    \multicolumn{5}{l}{\emph{Alt.\ backbone: Dinomaly + ResNet-18 base}} \\
    & bounded & \textbf{.972} & \textbf{.979} & --- \\
    & static mean & .947 & .967 & $-$2.50 \\
    & learned gate & .966 & .969 & $-$0.60 \\
    \bottomrule
  \end{tabular}
\end{table}

Table~\ref{tab:generalization} shows the results on both substitution settings. On the ResNet-18 backbone substitution, the deployed fusion rule outperforms static mean by +2.5pp I-AUROC and +1.2pp P-AUPRO, and also exceeds learned gate by +0.6pp I-AUROC and +1.0pp P-AUPRO. On the ResAD auxiliary substitution, the same rule outperforms static mean by +7.3pp I-AUROC and +2.7pp P-AUPRO, and also exceeds learned gate by +3.1pp I-AUROC and +10.8pp P-AUPRO. Across both settings, the same fusion rule consistently outperforms static pixel mean and learned gate. Absolute performance still varies with component strength, but the table suggests that the integration scheme remains useful under both auxiliary-branch and backbone substitution.

\paragraph{Hyperparameter sensitivity.}

A small hyperparameter sweep over $(\gamma, \alpha)$ under the default evaluation protocol is reported in Appendix~\ref*{app:supp-robustness}. I-AUROC changes within about 1.2pp across the tested grid and P-AUPRO within 0.30pp, suggesting that performance is not driven by narrow tuning.

\paragraph{Cross-dataset evidence.}

We further test TC-MAF on Eyecandies~\cite{bonfiglioli2022eyecandies}, a synthetic multimodal anomaly benchmark with aligned RGB and depth observations over 10 confectionery categories. We report the full 10-class mean under the same shared fusion formula and known-category calibration setup.

\begin{table}[t]
  \centering
  \caption{Results on Eyecandies. Results are 10-class means under the same shared fusion formula.}
  \label{tab:eyecandies}
  \begin{tabular}{lcc}
    \toprule
    Method & I-AUROC & P-AUPRO \\
    \midrule
    BTF'23~\cite{horwitz2023btf} & 0.821 & 0.846 \\
    M3DM'23~\cite{wang2023m3dm} & 0.897 & 0.877 \\
    CFM'24~\cite{costanzino2024cfmad} & 0.881 & 0.887 \\
    LDM'24~\cite{Liu2024LearningDM} & \second{0.948} & \second{0.941} \\
    3D-ADNAS'25~\cite{long2025revisiting} & 0.946 & 0.898 \\
    \midrule
    TC-MAF (Ours) & \best{0.951} & \best{0.954} \\
    \bottomrule
  \end{tabular}
\end{table}

Table~\ref{tab:eyecandies} reports the Eyecandies benchmark summary under the same shared fusion formula and known-category calibration setup. TC-MAF attains the highest I-AUROC and P-AUPRO in the table, showing that the same bounded fusion design transfers beyond MVTec-3D. This cross-dataset result complements the MVTec-3D benchmark lead and supports the method's robustness beyond one dataset.

\section{Conclusion}

We propose TC-MAF, a base-anchored multi-evidence fusion design for multimodal industrial anomaly detection. TC-MAF combines a multimodal detector, complementary Dinomaly evidence, and a small CMC term under one shared pixel-level fusion formula, while TDC provides a lightweight training-only calibration signal for scaling the auxiliary branch. On MVTec-3D, TC-MAF achieves 0.990 P-AUPRO and 0.979 I-AUROC, yielding the strongest mean benchmark results among the compared multimodal methods. Systematic ablations show that the fusion structure is the main source of improvement, while TDC contributes a smaller but reproducible gain and pooled-statistics calibration remains close to the default setting. Additional experiments show that the same design remains effective across auxiliary-branch and backbone substitutions, Eyecandies, few-shot evaluation, and a missing-3D setting.

\bibliographystyle{ACM-Reference-Format}
\bibliography{references}

\FloatBarrier

\clearpage
\newpage

\appendix
\section{Implementation Details}
\begin{figure}[t]
  \centering
  \includegraphics[width=\linewidth]{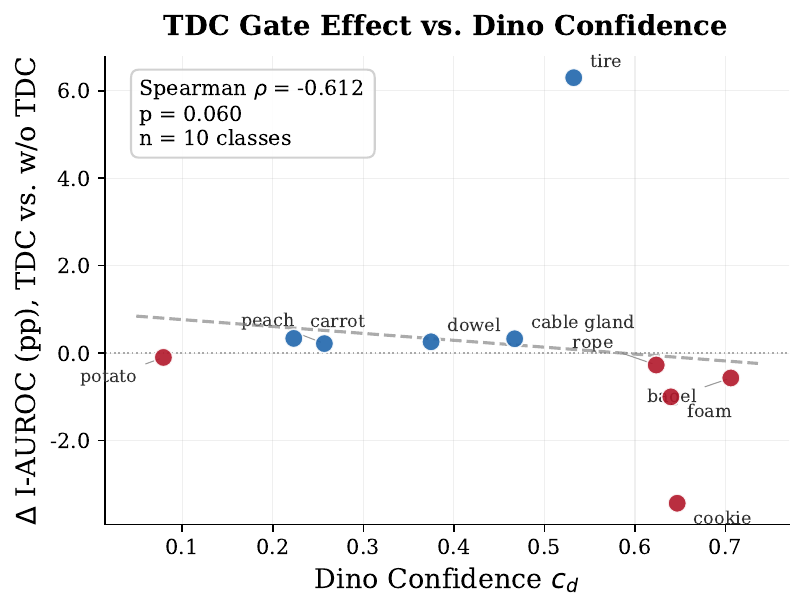}
  \caption{Direct class-wise view of the relation between TDC confidence and the signed TDC-specific image-level effect on MVTec-3D. Each point is one category. The mixed positive and negative effects indicate that TDC acts as a category-dependent calibration term rather than a uniform uplift.}
  \Description{Scatter plot showing classwise TDC confidence on the horizontal axis and the signed TDC-specific I-AUROC effect on the vertical axis, with class labels and a fitted trend line.}
  \label{fig:tdc-cd-gain}
\end{figure}
\begin{figure}[!htbp]
  \centering
  \includegraphics[width=0.95\columnwidth,height=0.88\textheight,keepaspectratio]{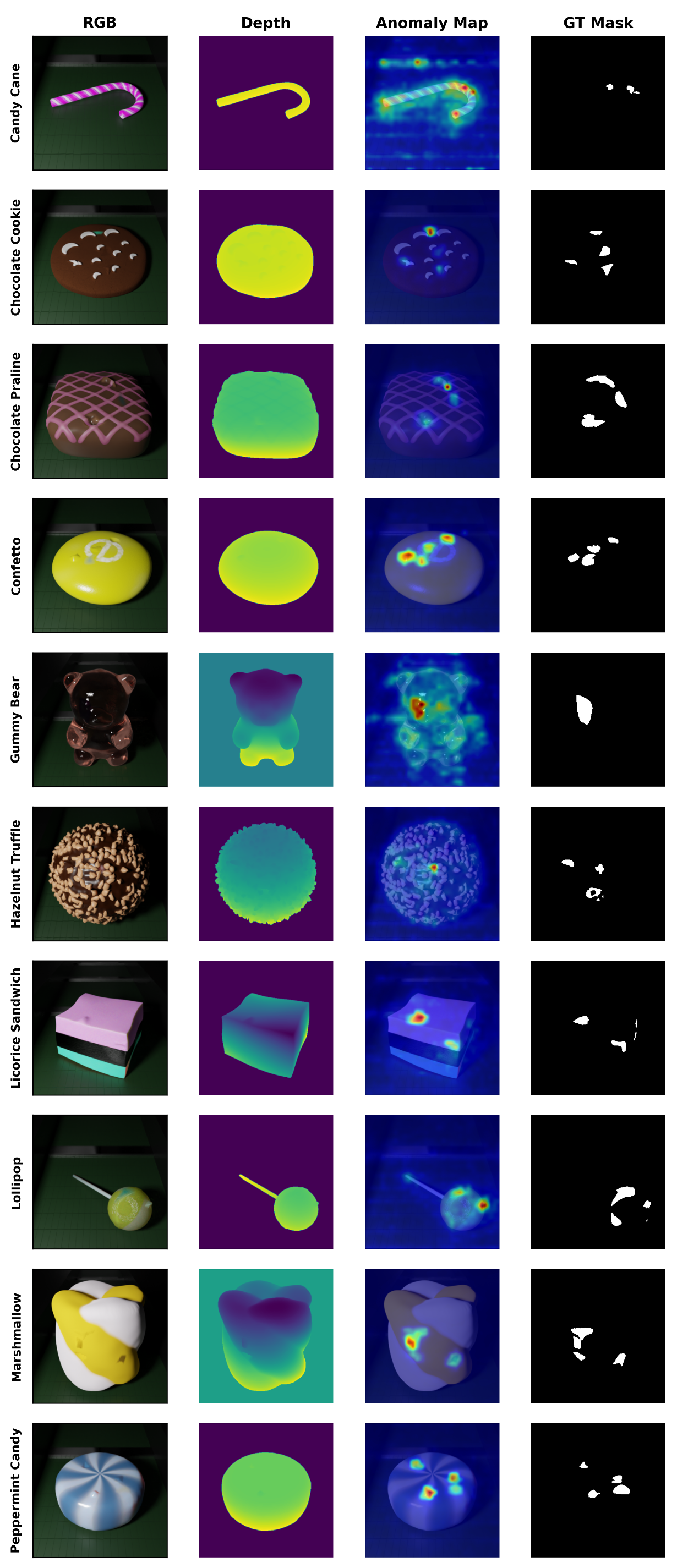}
  \caption{Full-class qualitative overview on Eyecandies. Each row shows one representative anomalous sample for one class with RGB, depth, the final TC-MAF anomaly map, and the ground-truth mask. Even when image-level separation is less pronounced on a few categories, the fused maps still highlight the annotated defect regions with locally concentrated responses.}
  \Description{Full-class qualitative overview on Eyecandies, with one row per class and four columns showing RGB, depth, TC-MAF anomaly map, and ground-truth mask.}
  \label{fig:appendix-eyecandies-fullclass}
\end{figure}
\begin{figure*}[!htbp]
  \centering
  \includegraphics[width=0.98\textwidth]{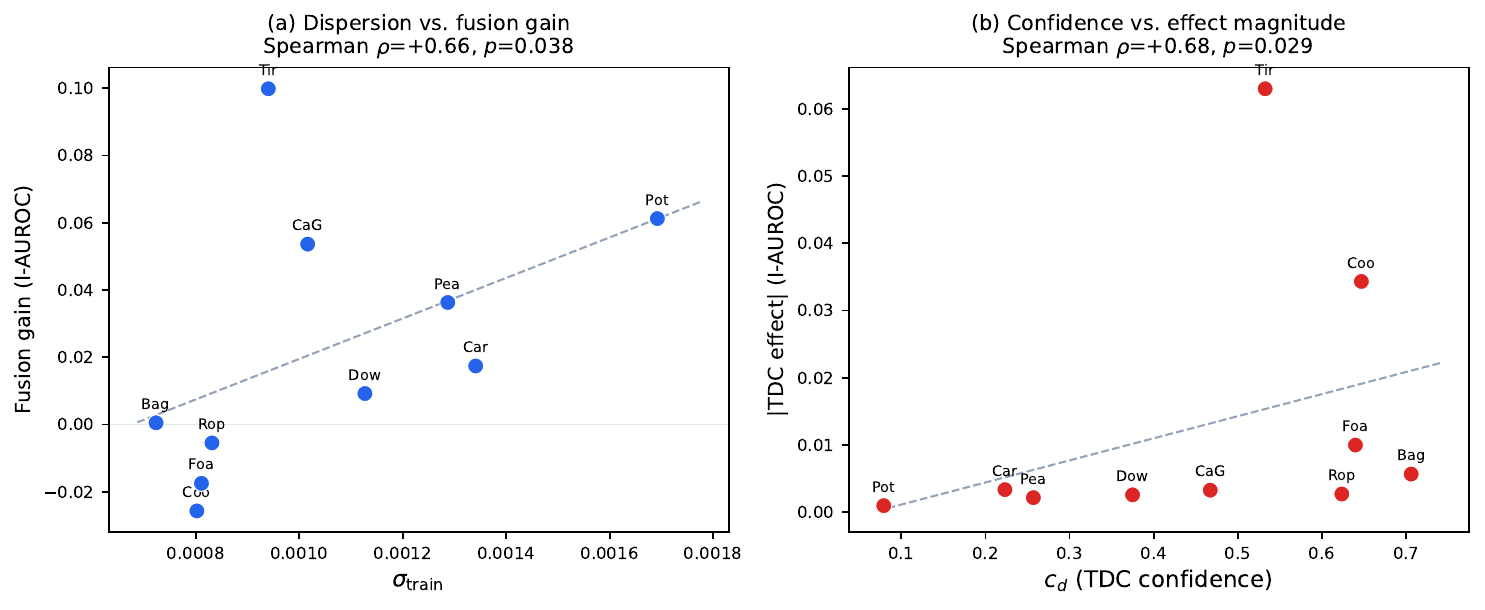}
  \caption{Per-class correlation between TDC inputs and observed fusion effects on MVTec-3D (10 categories). (a)~Training dispersion vs.\ fusion gain over the base-only branch. (b)~TDC confidence vs.\ absolute TDC-specific effect. The plots are consistent with interpreting TDC as a lightweight training-only calibration signal, while its average gain remains modest.}
  \Description{Scatter plots showing per-class correlation between training dispersion and fusion gain (left) and between TDC confidence and absolute TDC effect (right), with trend lines and Spearman statistics.}
  \label{fig:tdc-correlation}
\end{figure*}

\paragraph{Code entry point.}
The released pipeline is invoked via \path{run_tcmaf.sh}, which iterates over all 10 MVTec-3D classes and calls the evaluation script with the data root, checkpoint directories, GPU index, evaluation configuration, and output path. Cross-modal scoring logic comes from the shared fusion module to keep the implementation consistent with the base branch.

\paragraph{Dinomaly branch.}
The encoder follows Dinomaly~\cite{guo2025dinomaly} and uses a frozen DINOv2~\cite{oquab2024dinov2} ViT-S/14 backbone ($d{=}384$) with multi-layer extraction from target layers $\{2,3,\ldots,9\}$. Extracted features are grouped into two fusion sets of four layers each: $\{0,1,2,3\}$ and $\{4,5,6,7\}$ for both encoder and decoder sides. The decoder consists of 8 Transformer blocks ($d{=}384$, 6 heads, MLP ratio~4, linear attention) preceded by a single bottleneck MLP with dropout~0.2. Anomaly maps are computed from encoder--decoder cosine distance, smoothed with a Gaussian kernel ($k{=}5$, $\sigma{=}4$), and upsampled to input resolution. Input images are resized to $448{\times}448$, yielding $32{\times}32$ patch~tokens for the DINOv2 backbone.

\paragraph{Multimodal base branch.}
The base detector uses dual Wide-ResNet-50-2 backbones~\cite{zagoruyko2016wide} (frozen and ImageNet-pretrained) for RGB and geometry. Features are extracted from \texttt{layer2} and \texttt{layer3} and projected to 256~channels. A coreset memory bank is built from 10\% of training patches, and anomaly maps are produced by $k$-NN scoring ($k{=}3$) with z-normalization. Training runs for 20 epochs with Adam (lr~$10^{-4}$, weight decay~$10^{-5}$, batch size~4) on normal training data, with synthetic anomaly augmentation generated from those normal samples (probability~0.7, strength $[0.2, 0.8]$). Input resolution is $288{\times}288$ for both modalities.

\paragraph{Cross-modal mappers.}
Cross-modal reliability is computed by training lightweight residual MLP mappers between the normalized RGB and geometry patch features used by the deployed CMC module on the training set. The CMC map $M^{\mathrm{cross}}$ is the mean of bidirectional prediction errors (RGB$\to$Geo and Geo$\to$RGB), robustly normalized by the training-set median and scaled MAD.

\paragraph{Calibration constants.}
The two scalars in Eq.~\ref*{eq:dino-conf} are derived purely from training data: $\hat{\mu} = 0.000978$ and $\hat{\sigma} = 0.000291$, computed as the median and standard deviation of the category-specific Dinomaly training-score dispersions $\{\sigma_{\mathrm{train}}^{(c)}\}_{c=1}^{10}$ on MVTec-3D. These constants are fixed at deployment time and shared across all classes. In the default setting, each class uses its own $\sigma_{\mathrm{train}}^{(c)}$ together with these shared summary constants. Table~\ref*{tab:ablation-extended} shows that replacing the default per-class normalization statistics with pooled training statistics changes performance only marginally.

\section{Supplementary Sensitivity Results}\label{app:supp-robustness}

The missing-3D extended setting is already summarized in Group~D of Table~\ref*{tab:ablation-extended}; this section focuses on the supplementary hyperparameter sensitivity results.

\begin{table}[H]
  \centering
  \caption{Sensitivity to gate cap $\gamma$ and cross-modal weight $\alpha$. Bold marks the deployed default.}
  \label{tab:sensitivity}
  \small
  \begin{tabular}{c|ccc|ccc}
    \toprule
    & \multicolumn{3}{c|}{I-AUROC} & \multicolumn{3}{c}{P-AUPRO} \\
    $\gamma \backslash \alpha$ & 0.05 & 0.15 & 0.25 & 0.05 & 0.15 & 0.25 \\
    \midrule
    0.3 & .973 & .977 & .968 & .989 & .990 & .990 \\
    0.6 & .972 & \textbf{.979} & .976 & .990 & \textbf{.990} & .989 \\
    0.9 & .967 & .973 & .976 & .987 & .989 & .989 \\
    \bottomrule
  \end{tabular}
\end{table}

\paragraph{Fusion hyperparameters.}
All fusion equations and hyperparameters are fixed across the 10 classes: maximum gate $\gamma = 0.6$, CMC weight $\alpha = 0.15$, and disagreement threshold $\tau = 1.0$ (Eq.~\ref*{eq:gate-map}). For the final paper I-AUROC numbers, the image-level score is computed from stored fused maps using Eq.~\ref*{eq:topk-score}; no fusion map is regenerated for this step.

\section{Metric Fairness and Qualitative Analysis}\label{app:classwise-results}

\paragraph{Comparison protocol note.}
All benchmark scores reported here use the same metric implementations: image-level AUROC and pixel-level AUROC are computed with \texttt{roc\_auc\_score}, and P-AUPRO uses the standard PRO pipeline (200 thresholds, $5\times5$ dilation, FPR$<0.3$, area-under-curve after FPR normalization). This appendix summarizes the metric setup used in our final presentation.

\paragraph{Efficiency breakdown.}
Dinomaly requires 15.7$\pm$0.1\,ms and 240\,MB, while the multimodal base branch requires 43.9$\pm$2.3\,ms and 1564\,MB. Sequential execution totals 59.6\,ms and 1804\,MB, so runtime is dominated by evidence generation rather than the lightweight adaptive fusion itself.

\begin{table}[H]
  \centering
  \caption{Efficiency breakdown showing branch-wise cost composition.}
  \label{tab:efficiency}
  \begin{tabular}{lccc}
    \toprule
    Method & Mem.\ (MB) & FPS & Latency (ms) \\
    \midrule
    Shape-Guided'23~\cite{chu2023shapeguided} & 6155 & 0.3 & 3333.3 \\
    M3DM'23~\cite{wang2023m3dm} & 11962 & 0.7 & 1428.6 \\
    CFM'24~\cite{costanzino2024cfmad} & 621 & 21 & 47.6 \\
    3DSR'24~\cite{zavrtanik2024cheating} & 2921 & 38 & 26.3 \\
    \midrule
    Dinomaly branch & 240 & 63.9 & 15.7$\pm$0.1 \\
    Multimodal base branch & 1564 & 22.8 & 43.9$\pm$2.3 \\
    TC-MAF (Ours) & 1804 & 16.8 & 59.6 \\
    \bottomrule
  \end{tabular}
\end{table}

\section{Few-Shot Class-Wise Results}\label{app:fewshot-classwise}

This section provides the per-class breakdown of few-shot experiments on MVTec-3D. Tables~\ref{tab:fewshot-tcmaf-iauroc} and~\ref{tab:fewshot-tcmaf-paupro} show TC-MAF's results for 1/5/10/50-shot together with the full-shot reference.

\begin{table*}[!htbp]
  \centering
  \caption{TC-MAF class-wise I-AUROC on MVTec-3D few-shot settings.}
  \label{tab:fewshot-tcmaf-iauroc}
  \begin{tabular}{l|cccccccccc|c}
    \toprule
    Setting & Bagel & Cable Gland & Carrot & Cookie & Dowel & Foam & Peach & Potato & Rope & Tire & Mean \\
    \midrule
    1-shot & 0.574 & 0.524 & 0.589 & 0.801 & 0.627 & 0.797 & 0.573 & 0.798 & 0.741 & 0.631 & 0.666 \\
    5-shot & 0.886 & 0.641 & 0.799 & 0.793 & 0.871 & 0.851 & 0.845 & 0.830 & 0.953 & 0.832 & 0.830 \\
    10-shot & 0.936 & 0.779 & 0.813 & 0.830 & 0.922 & 0.864 & 0.883 & 0.876 & 0.960 & 0.864 & 0.873 \\
    50-shot & 0.977 & 0.991 & 0.970 & 0.882 & 0.967 & 0.897 & 0.951 & 0.903 & 0.963 & 0.912 & 0.941 \\
    full & 0.992 & 0.998 & 0.992 & 0.953 & 0.992 & 0.952 & 0.988 & 0.979 & 0.984 & 0.958 & 0.979 \\
    \bottomrule
  \end{tabular}
\end{table*}

\begin{table*}[!htbp]
  \centering
  \caption{TC-MAF class-wise P-AUPRO on MVTec-3D few-shot settings.}
  \label{tab:fewshot-tcmaf-paupro}
  \begin{tabular}{l|cccccccccc|c}
    \toprule
    Setting & Bagel & Cable Gland & Carrot & Cookie & Dowel & Foam & Peach & Potato & Rope & Tire & Mean \\
    \midrule
    1-shot & 0.819 & 0.715 & 0.895 & 0.824 & 0.821 & 0.902 & 0.874 & 0.952 & 0.891 & 0.832 & 0.853 \\
    5-shot & 0.987 & 0.857 & 0.975 & 0.951 & 0.979 & 0.893 & 0.990 & 0.977 & 0.928 & 0.985 & 0.952 \\
    10-shot & 0.973 & 0.955 & 0.982 & 0.963 & 0.976 & 0.883 & 0.992 & 0.985 & 0.933 & 0.981 & 0.962 \\
    50-shot & 0.989 & 0.967 & 0.986 & 0.981 & 0.984 & 0.905 & 0.994 & 0.992 & 0.953 & 0.983 & 0.973 \\
    full & 0.993 & 0.997 & 0.997 & 0.989 & 0.991 & 0.979 & 0.998 & 0.998 & 0.963 & 0.992 & 0.990 \\
    \bottomrule
  \end{tabular}
\end{table*}

\section{Cross-Dataset Class-Wise Results on Eyecandies}\label{app:eyecandies-classwise}

Tables~\ref{tab:eyecandies-classwise-iauroc} and~\ref{tab:eyecandies-classwise-paupro} report the per-category I-AUROC and P-AUPRO comparison on Eyecandies, complementing the mean-level summary in Table~\ref*{tab:eyecandies}. TC-MAF uses the same shared fusion formula and known-category calibration setup as on MVTec-3D.

\begin{table*}[!htbp]
  \centering
  \caption{Class-wise I-AUROC comparison on Eyecandies. TC-MAF uses the same fixed fusion rule as on MVTec-3D.}
  \label{tab:eyecandies-classwise-iauroc}
  \resizebox{\textwidth}{!}{
  \begin{tabular}{lcccccccccccc}
    \toprule
    Method & Candy C. & Choc.\ Cook. & Choc.\ Pral. & Confetto & Gummy B. & Hazel.\ T. & Licor.\ S. & Lollipop & Marshm. & Pepper.\ C. & Mean \\
    \midrule
    BTF'23~\cite{horwitz2023btf} & 0.712 & 0.882 & 0.784 & 0.942 & 0.774 & 0.579 & 0.829 & 0.840 & 0.986 & 0.882 & 0.821 \\
    EasyNet'23~\cite{chen2023easynet} & 0.737 & 0.934 & 0.866 & 0.966 & 0.717 & 0.822 & 0.847 & 0.863 & 0.977 & 0.960 & 0.869 \\
    M3DM'23~\cite{wang2023m3dm} & 0.624 & 0.958 & 0.958 & \textbf{1.000} & 0.886 & 0.758 & 0.949 & 0.836 & \textbf{1.000} & \textbf{1.000} & 0.897 \\
    3DSR'24~\cite{zavrtanik2024cheating} & 0.651 & 0.998 & 0.904 & 0.978 & 0.875 & 0.861 & 0.965 & 0.899 & 0.990 & 0.971 & 0.909 \\
    CFM'24~\cite{costanzino2024cfmad} & 0.680 & 0.931 & 0.952 & 0.880 & 0.865 & 0.782 & 0.917 & 0.840 & 0.998 & 0.962 & 0.881 \\
    LDM'24~\cite{Liu2024LearningDM} & 0.859 & \textbf{1.000} & \textbf{1.000} & 0.995 & \textbf{0.910} & 0.738 & \textbf{0.998} & \textbf{0.976} & \textbf{1.000} & \textbf{1.000} & 0.948 \\
    3D-ADNAS'25~\cite{long2025revisiting} & \textbf{0.896} & \textbf{1.000} & 0.970 & \textbf{1.000} & 0.827 & \textbf{0.882} & 0.931 & 0.950 & \textbf{1.000} & \textbf{1.000} & 0.946 \\
    \midrule
    TC-MAF (Ours) & \textbf{0.896} & 0.998 & 0.989 & 0.998 & 0.905 & 0.841 & 0.952 & 0.938 & \textbf{1.000} & 0.988 & \textbf{0.951} \\
    \bottomrule
  \end{tabular}
  }
\end{table*}

\begin{table*}[!htbp]
  \centering
  \caption{Per-class P-AUROC under full-shot setting. TC-MAF uses the same fixed fusion rule across all classes.}
  \label{tab:pauroc-fullshot-perclass}
  \small
  \setlength{\tabcolsep}{5pt}
  \renewcommand{\arraystretch}{1.15}
  \begin{tabular}{ccccccccccc}
    \toprule
    \multicolumn{11}{c}{\textbf{MVTec-3D }} \\
    \midrule
    Bagel & Cable G. & Carrot & Cookie & Dowel & Foam & Peach & Potato & Rope & Tire & Mean \\
    \midrule
    0.997 & 0.998 & 0.999 & 0.996 & 0.998 & 0.992 & 0.999 & 0.999 & 0.988 & 0.998 & \textbf{0.996} \\
    \midrule
    \multicolumn{11}{c}{\textbf{Eyecandies }} \\
    \midrule
    Candy C. & Choc.\ Cook. & Choc.\ Pral. & Confetto & Gummy B. & Hazel.\ T. & Licor.\ S. & Lollipop & Marshm. & Pepper.\ C. & Mean \\
    \midrule
    0.988 & 0.993 & 0.983 & 0.989 & 0.976 & 0.965 & 0.986 & 0.991 & 0.989 & 0.994 & \textbf{0.985} \\
    \bottomrule
  \end{tabular}
\end{table*}
\begin{table*}[!htbp]
  \centering
  \caption{Class-wise P-AUPRO comparison on Eyecandies. TC-MAF uses the same fixed fusion rule as on MVTec-3D.}
  \label{tab:eyecandies-classwise-paupro}
  \resizebox{\textwidth}{!}{
  \begin{tabular}{lcccccccccccc}
    \toprule
    Method & Candy C. & Choc.\ Cook. & Choc.\ Pral. & Confetto & Gummy B. & Hazel.\ T. & Licor.\ S. & Lollipop & Marshm. & Pepper.\ C. & Mean \\
    \midrule
    BTF'23~\cite{horwitz2023btf} & 0.871 & 0.900 & 0.698 & 0.966 & 0.823 & 0.567 & 0.884 & 0.905 & 0.953 & 0.897 & 0.846 \\
    M3DM'23~\cite{wang2023m3dm} & 0.906 & 0.826 & 0.803 & \textbf{0.983} & 0.855 & 0.688 & 0.880 & 0.906 & 0.966 & 0.955 & 0.877 \\
    CFM'24~\cite{costanzino2024cfmad} & 0.942 & 0.902 & 0.831 & 0.965 & 0.875 & 0.762 & 0.791 & 0.913 & 0.939 & 0.949 & 0.887 \\
    LDM'24~\cite{Liu2024LearningDM} & 0.964 & 0.953 & 0.951 & 0.982 & 0.931 & 0.765 & \textbf{0.969} & 0.935 & \textbf{0.982} & 0.983 & 0.941 \\
    3D-ADNAS'25~\cite{long2025revisiting} & 0.945 & 0.891 & 0.827 & 0.958 & 0.857 & 0.748 & 0.911 & 0.907 & 0.964 & 0.972 & 0.898 \\
    \midrule
    TC-MAF (Ours) & \textbf{0.972} & \textbf{0.969} & \textbf{0.967} & 0.980 & \textbf{0.948} & \textbf{0.824} & 0.959 & \textbf{0.971} & 0.965 & \textbf{0.984} & \textbf{0.954} \\
    \bottomrule
  \end{tabular}
  }
\end{table*}
\begin{table}[H]
  \centering
  \caption{Mean-level P-AUROC comparison on full-shot setting. TC-MAF uses the same fixed fusion rule across all classes.}
  \label{tab:appendix-pauroc}
  \begin{tabular}{lc}
    \toprule
    Method & MVTec-3D  \\
    \midrule
    BTF'23~\cite{horwitz2023btf} & 0.992 \\
    AST'23~\cite{rudolph2023ast} & 0.976  \\
    M3DM'23~\cite{wang2023m3dm} & 0.992  \\
    CFM'24~\cite{costanzino2024cfmad} & 0.993 \\
    LSFA'24~\cite{tu2024lsfa} & 0.993  \\
    \midrule
    TC-MAF (Ours) & \textbf{0.996}  \\
    \bottomrule
  \end{tabular}
\end{table}
Among the methods summarized in Table~\ref{tab:appendix-pauroc}, TC-MAF attains the highest mean P-AUROC on MVTec-3D. Table~\ref{tab:appendix-pauroc} summarizes that mean-level comparison, while Table~\ref{tab:pauroc-fullshot-perclass} provides the per-class TC-MAF P-AUROC breakdown for MVTec-3D and Eyecandies. Together with the Eyecandies mean P-AUROC of 0.985 in Table~\ref{tab:pauroc-fullshot-perclass}, these results show that TC-MAF's localization gains are also reflected in pixel-level classification accuracy.

\section{Evidence Source Ablation}\label{app:modality-ablation}

Table~\ref{tab:modality-ablation} decomposes TC-MAF into its constituent evidence sources on MVTec-3D. ``Dinomaly (RGB recon.)'' is the standalone reconstruction branch; ``Multimodal base'' is the multimodal base detector that jointly processes RGB and 3D inputs; ``TC-MAF'' is the full adaptive fusion. Under this internal evaluation protocol, the fused result is higher than either individual source on all three reported metrics, supporting the view that TC-MAF combines complementary evidence within the tested setup.

\begin{table}[H]
  \centering
  \caption{Evidence source decomposition on MVTec-3D. Each row uses a single evidence source or the full TC-MAF fusion. All numbers use the same evaluation protocol.}
  \label{tab:modality-ablation}
  \small
  \begin{tabular}{lccc}
    \toprule
    Evidence source & I-AUROC & P-AUROC & P-AUPRO \\
    \midrule
    Dinomaly (RGB recon.) & 0.924 & 0.992 & 0.977 \\
    Multimodal base & 0.955 & 0.992 & 0.981 \\
    \midrule
    TC-MAF (Ours) & \textbf{0.979} & \textbf{0.996} & \textbf{0.990} \\
    \bottomrule
  \end{tabular}
\end{table}

Notably, the multimodal base alone already achieves strong P-AUPRO (0.981), and TC-MAF's adaptive fusion with Dinomaly lifts it to 0.990 (+0.9pp). The I-AUROC gain from fusion is larger (+2.3pp over the base), showing that the reconstruction branch contributes complementary evidence that the gate mechanism successfully harvests.

\section{Alternative Proxy Comparison}\label{app:proxy-comparison}

Table~\ref{tab:proxy-comparison} reports proxy comparison results on MVTec-3D. Here, ``w/o TDC'' removes the training-derived scaling term from the deployed gate, whereas the other rows retain the same gate structure and only change the proxy used to instantiate $c_d$. Std, MAD, and mean absolute deviation remain close, whereas entropy-like and residual-dispersion proxies are weaker.

\begin{table}[H]
  \centering
  \caption{Proxy comparison for training-only calibration signals on MVTec-3D. Std, MAD, and mean absolute deviation proxies belong to the same train-dispersion family and remain close; entropy-like and residual-dispersion proxies are weaker.}
  \label{tab:proxy-comparison}
  \begin{tabular}{lcc}
    \toprule
    Proxy & I-AUROC & P-AUPRO \\
    \midrule
    Std-based TDC (ours) & \textbf{0.9788} & 0.9901 \\
    MAD-like & 0.9786 & 0.9901 \\
    Mean abs.\ deviation & 0.9784 & \textbf{0.9909} \\
    Residual dispersion & 0.9661 & 0.9896 \\
    Entropy-like & 0.9664 & 0.9865 \\
    w/o TDC & 0.9709 & 0.9881 \\
    \bottomrule
  \end{tabular}
\end{table}

\section{TDC Correlation Details}\label{app:tdc-correlation}

This section collects the per-class diagnostics behind the main-paper discussion of TDC. Figure~\ref{fig:tdc-cd-gain} plots the signed TDC-specific image-level effect against TDC confidence for the 10 MVTec-3D categories. The mixed positive and negative signed effects are consistent with interpreting TDC as a category-dependent calibration term under the shared fusion rule rather than as a uniform uplift. Figure~\ref{fig:tdc-correlation} complements this view with the absolute-effect analysis used in the corresponding Spearman tests.

\section{Pixel-Level AUROC Results}\label{app:pauroc}

Table~\ref{tab:appendix-pauroc} reports mean-level pixel-level AUROC (P-AUROC) for MVTec-3D. TC-MAF numbers are computed from the same shared fusion formula under the benchmark protocol used throughout this paper.

For completeness, TC-MAF also achieves 0.985 P-AUROC on Eyecandies, and Table~\ref{tab:pauroc-fullshot-perclass} provides the per-class breakdown. Together with the MVTec-3D P-AUROC of 0.996 (Table~\ref{tab:appendix-pauroc}), this shows that TC-MAF's localization strength extends to pixel-level classification accuracy, not only region-level precision (P-AUPRO).

Table~\ref{tab:pauroc-fullshot-perclass} provides the full per-class breakdown for both datasets.

\end{document}